\documentclass[11pt,a4paper]{article}
\usepackage[hyperref]{acl2021}
\usepackage{times}
\usepackage{latexsym}

\usepackage{microtype}

\usepackage{graphicx}  
\usepackage{xspace}
\usepackage{eufrak}
\usepackage{amsfonts}
\usepackage{amssymb}
\usepackage{amsmath}
\usepackage{bm}
\usepackage{multirow}
\usepackage{booktabs}
\usepackage{subfig}
\usepackage[utf8]{inputenc}

\usepackage{cleveref}
\usepackage{enumitem}

\usepackage{xargs}

\usepackage[colorinlistoftodos,prependcaption,textsize=tiny]{todonotes}
\usepackage[group-separator={,}, group-minimum-digits=4]{siunitx}

\newcommand{\ours}{{CHMM-ALT}\xspace}
\newcommand{\generative}{{CHMM}\xspace}
\newcommand*{\rom}[1]{\uppercase\expandafter{\romannumeral #1}}

\newcommandx{\chao}[2][1=]{\todo[linecolor=red,backgroundcolor=red!25,bordercolor=red,#1]{#2}}
\newcommandx{\yli}[2][1=]{\todo[linecolor=blue,backgroundcolor=blue!25,bordercolor=blue,#1]{#2}}
\newcommandx{\lsong}[2][1=]{\todo[linecolor=green,backgroundcolor=green!25,bordercolor=green,#1]{#2}}
\newcommandx{\pranav}[2][1=]{\todo[linecolor=purple,backgroundcolor=purple!25,bordercolor=red,#1]{#2}}
\newcommandx{\improvement}[2][1=]{\todo[linecolor=yellow,backgroundcolor=yellow!25,bordercolor=yellow,#1]{#2}}
\newcommandx{\thiswillnotshow}[2][1=]{\todo[disable,#1]{#2}}

\crefname{section}{§}{§§}
\Crefname{section}{§}{§§}

\DeclareMathOperator*{\argmax}{arg\,max}
\DeclareMathOperator*{\argmin}{arg\,min}

\newcommand{\ie}{\emph{i.e.}\xspace} 
\newcommand{\eg}{\emph{e.g.}\xspace} 
\newcommand{\wrt}{\emph{w.r.t.}\xspace}

\newcommand{\x}{\bm{x}}
\newcommand{\0}{\bm{0}}
\newcommand{\y}{\bm{y}}

\newcommand{\z}{\bm{z}}
\newcommand{\e}{\bm{e}}
\newcommand{\w}{\bm{w}}

\newcommand{\lbs}{\mathcal{L}}
\newcommand{\R}{\mathbb{R}}

\newcommand{\E}{\mathcal{E}}

\newcommand{\bPsi}{\bm{\Psi}}
\newcommand{\bPhi}{\bm{\Phi}}
\newcommand{\bphi}{\bm{\varphi}}
\newcommand{\ba}{\bm{\alpha}}
\newcommand{\bpi}{\bm{\pi}}
\newcommand{\btht}{\bm{\theta}}

\newcommand{\kld}{\mathcal{D}}

\newcommand{\expt}{\mathbb{E}}

\aclfinalcopy 

\title{BERTifying the Hidden Markov Model for Multi-Source Weakly Supervised Named Entity Recognition}

\author{
  Yinghao Li\textsuperscript{1},
  Pranav Shetty\textsuperscript{1},
  Lucas Liu\textsuperscript{1},
  Chao Zhang\textsuperscript{1},
  \and Le Song\textsuperscript{2} \\
\textsuperscript{1} Georgia Institute of Technology, Atlanta, USA \\
\textsuperscript{2} Mohamed bin Zayed University of Artificial Intelligence, Abu Dhabi, United Arab Emirates \\
\texttt{ \{yinghaoli, pranav.shetty, lucasliu, chaozhang\}@gatech.edu } \\
\texttt{ le.song@mbzuai.ac.ae } }

\date{}

\begin{document}
\maketitle

\begin{abstract}
We study the problem of learning a named entity recognition (NER) tagger using noisy labels from multiple weak supervision sources.
Though cheap to obtain, the labels from weak supervision sources are often incomplete, inaccurate, and contradictory, making it difficult to learn an accurate NER model.
To address this challenge, we propose a conditional hidden Markov model (\generative), which can effectively infer true labels from multi-source noisy labels in an unsupervised way.
\generative enhances the classic hidden Markov model with the contextual representation power of pre-trained language
models.
Specifically, \generative learns token-wise transition and emission probabilities from the BERT embeddings of the input tokens to infer the latent true labels from noisy observations.
We further refine \generative with an alternate-training approach (\ours).
It fine-tunes a BERT-NER model with the labels inferred by \generative, and this BERT-NER's output is regarded as an additional weak source to train the \generative in return.
Experiments on four NER benchmarks from various domains show that our method outperforms state-of-the-art weakly supervised NER models by wide margins.
\end{abstract}

\section{Introduction}
\label{sec:introduction}

Named entity recognition (NER), which aims to identify named entities from unstructured text, is an information extraction task fundamental to many downstream applications such as event detection \citep{Li2012twevent}, relationship extraction \citep{bach2007review}, and question answering \citep{Khalid-2008-the-impact}.
Existing NER models are typically supervised by a large number of training sequences, each pre-annotated with token-level labels.
In practice, however, obtaining such labels could be prohibitively expensive.
On the other hand, many domains have various knowledge resources such as knowledge bases, domain-specific dictionaries, or labeling rules provided by domain experts \citep{Farmakiotou-2000-rule-basednamed, Nadeau-2007-a-survey}.
These resources can be used to match a corpus and quickly create large-scale noisy training data for NER from multiple views.

Learning an NER model from multiple weak supervision sources is a challenging problem.
While there are works on distantly supervised NER that use only knowledge bases as weak supervision \citep{mintz-etal-2009-distant, shang-etal-2018-learning, cao-etal-2019-low, Liang-etal-2020-bond}, they cannot leverage complementary information from multiple annotation sources.
To handle multi-source weak supervision, several recent works \citep{nguyen-etal-2017-aggregating,Safranchik-etal-2020-weakly,lison-etal-2020-named} leverage the hidden Markov model (HMM), by modeling true labels as hidden variables and inferring them from the observed noisy labels through unsupervised learning.
Though principled, these models fall short in capturing token semantics and context information, as they either model input tokens as one-hot observations \citep{nguyen-etal-2017-aggregating} or do not model them at all \citep{Safranchik-etal-2020-weakly, lison-etal-2020-named}.
Moreover, the flexibility of HMM is limited as its transitions and emissions remain constant over time steps, whereas in practice they should depend on the input words.

We propose the conditional hidden Markov model (\generative) to infer true NER labels from multi-source weak annotations.
\generative conditions the HMM training and inference on BERT by predicting token-wise transition and emission probabilities from the BERT embeddings.
These token-wise probabilities are more flexible than HMM's constant counterpart in modeling how the true labels should evolve according to the input tokens.
The context representation ability they inherit from BERT also relieves the Markov constraint and expands HMM's context-awareness.

Further, we integrate \generative with a supervised BERT-based NER mode with an alternate-training method (\ours).
It fine-tunes BERT-NER with the denoised labels generated by \generative.
Taking advantage of the pre-trained knowledge contained in BERT, this process aims to refine the denoised labels by discovering the entity patterns neglected by all of the weak sources.
The fine-tuned BERT-NER serves as an additional supervision source, whose output is combined with other weak labels for the next round of \generative training.
\ours trains \generative and BERT-NER alternately until the result is optimized.

Our contributions include:
\begin{itemize}
    \item A multi-source label aggregator \generative with token-wise transition and emission probabilities for aggregating multiple sets of NER labels from different weak labeling sources.
    \item An alternate-training method \ours that trains \generative and BERT-NER in turn utilizing each other's outputs for multiple loops to optimize the multi-source weakly supervised NER performance.
    \item A comprehensive evaluation on four NER benchmarks from different domains demonstrates that \ours achieves a $4.83$ average F1 score improvement over the strongest baseline models.
\end{itemize}
The code and data used in this work are available at \href{https://github.com/Yinghao-Li/CHMM-ALT}{github.com/Yinghao-Li/CHMM-ALT}.

\section{Related Work}
\label{sec:related.works}

\paragraph{Weakly Supervised NER}

There have been works that train NER models with different weak supervision approaches.
\textit{Distant supervision}, a specific type of weak supervision, generates training labels from knowledge bases \citep{mintz-etal-2009-distant,yang-etal-2018-distantly, shang-etal-2018-learning, cao-etal-2019-low, Liang-etal-2020-bond}.
But such a method is limited to one source and falls short of acquiring supplementary annotations from other available resources.
Other works adopt multiple additional labeling sources, such as heuristic functions that depend on lexical features, word patterns, or document information \citep{Nadeau-2007-a-survey, Ratner-2016-data-programming}, and unify their results through multi-source \textit{label denoising}.
Several multi-source weakly supervised learning approaches are designed for sentence classification \citep{Ratner-2017-Snorkel, Ratner-2019-training-complex, ren-etal-2020-denoising, yu2020finetuning}.
Although these methods can be adapted for sequence labeling tasks such as NER, they tend to overlook the internal dependency relationship between token-level labels during the inference.
\citet{Fries-2017-SwellShark} target the NER task, but their method first generates candidate named entity spans and then classifies each span independently.
This independence makes it suffer from the same drawback as sentence classification models.

A few works consider label dependency while dealing with multiple supervision sources.
\citet{lan-etal-2020-learning} train a BiLSTM-CRF network \citep{huang2015bidirectional} with multiple parallel CRF layers, each for an individual labeling source, and aggregate their transitions with confidence scores predicted by an attention network \citep{Bahdanau-2015-neural, luong-etal-2015-effective}.
HMM is a more principled model for multi-source sequential label denoising as the true labels are implicitly inferred through unsupervised learning without deliberately assigning any additional scores.
Following this track, \citet{nguyen-etal-2017-aggregating} and \citet{lison-etal-2020-named} use a standard HMM with multiple observed variables, each from one labeling source.
\citet{Safranchik-etal-2020-weakly} propose linked HMM, which differs from ordinary HMM by introducing unique linking rules as an adjunct supervision source additional to general token labels.
However, these methods fail to utilize the context information embedded in the tokens as effectively as \generative, and their NER  performance is further constrained by the Markov assumption.

\paragraph{Neuralizing the Hidden Markov Model}

Some works attempt to neuralize HMM in order to relax the Markov assumption while maintaining its generative property \citep{Kim-2018-a-Tutorial}.
For example, \citet{Dai-2017-recurrent} and \citet{Liu-2018-structured-inference} incorporate recurrent units into the hidden semi-Markov model (HSMM) to segment and label high-dimensional time series;
\citet{wiseman-etal-2018-learning} learn discrete template structures for conditional text generation using neuralized HSMM.
\citet{Wessels-2000-refining} and \citet{chiu-rush-2020-scaling} factorize HMM with neural networks to scale it and improve its sequence modeling capacity.
The work most related to ours leverages neural HMM for sequence labeling \citep{tran-etal-2016-unsupervised}.
\generative differs from neural HMM in that the tokens are treated as a dependency term in \generative instead of the observation in neural HMM.
Besides, \generative is trained with generalized EM, whereas neural HMM optimizes the marginal likelihood of the observations.

\begin{figure}[!t]
    \centerline{\includegraphics[width = 0.5\textwidth]{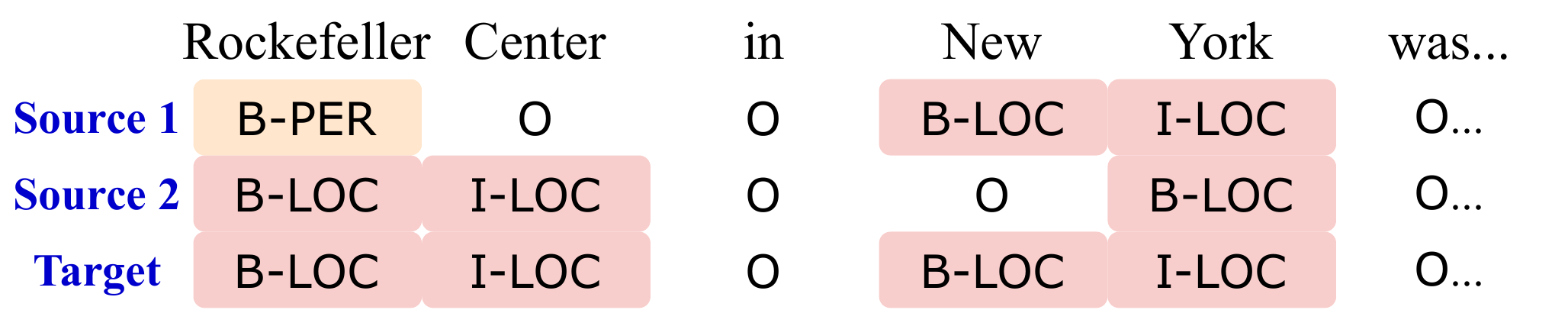}}
    \caption{
        An example of label aggregation with two weak labeling sources.
        We use BIO labeling scheme.
         \texttt{PER} represents person; \texttt{LOC} is location.
    }
    \label{fig:1.label.aggregation}
\end{figure}

\begin{figure*}[tbp]
    \centerline{\includegraphics[width = \textwidth]{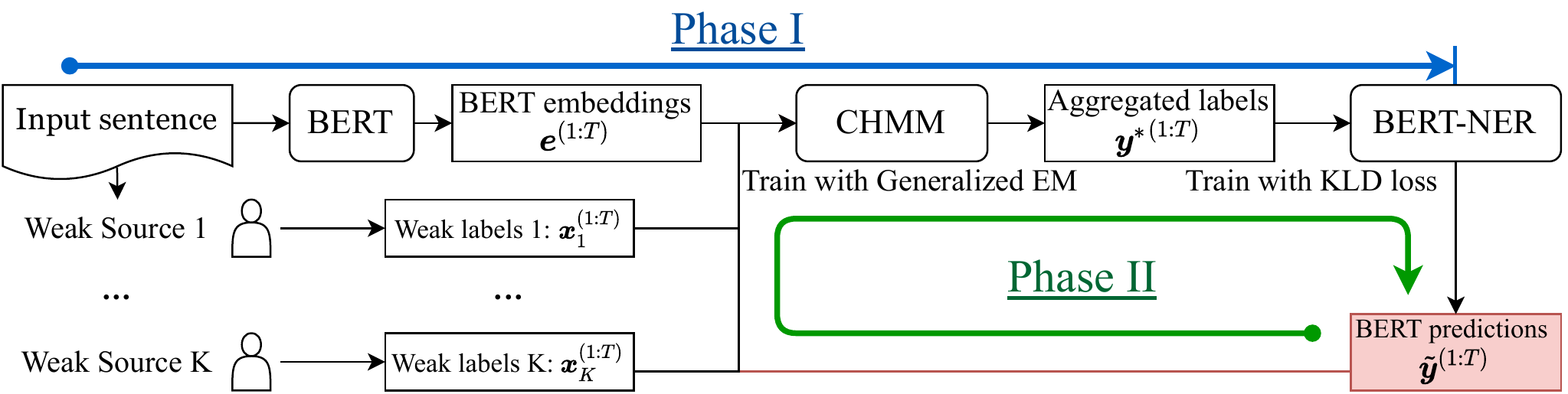}}
    \caption{
        The illustration of the alternate-training method.
        Phase~\rom{1} is acyclic, starting from getting $K$ weak labels from supervision sources and ending at the fine-tuning of BERT-NER with \generative's denoised output.
        Phase~\rom{2} contains several loops, each trains \generative with $K+1$ sources, including the additional BERT predictions from the previous loop, and fine-tunes BERT-NER using the updated denoised labels.
    }
    \label{fig:3.cross.training}
\end{figure*}

\section{Problem Setup}
\label{sec:problem.setup}

In this section, we formulate the multi-source weakly supervised NER problem.
Consider an input sentence that contains $T$ tokens $\w^{(1:T)}$, NER can be formulated as a sequence labeling task that assigns a label to each token in the sentence.\footnote{We represent vectors, matrices or tensors with bold fonts and scalars with regular fonts; $1:a \triangleq \{1,2,\dots,a\}$.}
Assuming the set of target entity types is $\E$ and the tagging scheme is BIO \citep{ramshaw-marcus-1995-text}, NER models assign one label from the label set $l \in \lbs$ to each token, where the size of the label set is $|\lbs| = 2|\E|+1$, \eg, if $\E = \{\texttt{PER}, \texttt{LOC}\}$, then $\lbs = \{ \texttt{O}, \texttt{B-PER}, \texttt{I-PER}, \texttt{B-LOC}, \texttt{I-LOC} \}$.

Suppose we have a sequence with $K$ weak sources, each of which can be a heuristic rule, knowledge base, or existing out-of-domain NER model.
Each source serves as a labeling function that generates token-level weak labels from the input corpus, as shown in Figure~\ref{fig:1.label.aggregation}.
For the input sequence $\w^{(1:T)}$, we use $\x^{(1:T)}_k, k\in\{1,\dots, K\}$ to represent the weak labels from the source $k$,
where $\x^{(t)}_k \in \R^{|\lbs|}, t \in \{1,\dots, T\}$ is a probability distribution over $\lbs$.
Multi-source weakly supervised NER aims to find the underlying true sequence of labels $\hat{\y}^{(1:T)}, \hat{y}^{(t)}\in \lbs$ given $\{\w^{(1:T)}, \x^{(1:T)}_{1:K} \}$.

\section{Methodology}
\label{sec:methodology}

In this section, we describe our proposed method \ours.
We first sketch the alternate-training procedure (\cref{subsec:method.cross.training}), then explain the \generative component (\cref{subsec:method.nhmm}) and how BERT-NER is involved (\cref{subsec:bert.ner}).

\subsection{Alternate-Training Procedure}
\label{subsec:method.cross.training}

The alternate-training method trains two models---a multi-source label aggregator \generative and a BERT-NER model---in turn with each other's output.
\generative aggregates multiple sets of labels from different sources into a unified sequence of labels, while BERT-NER refines them by its language modeling ability gained from pre-training.
The training process is divided into two phases.

\begin{itemize}
\item In \textbf{phase~\rom{1}}, \generative takes the annotations $\x^{(1:T)}_{1:K}$ from existing sources and gives a set of denoised labels ${\y^*}^{(1:T)}$, which are used to fine-tune the BERT-NER model.
Then, we regard the fine-tuned model as an additional labeling source, whose outputs ${\tilde{\y}}^{(1:T)}$ are added into the original weak label sets to give the updated observation instances: $\x^{(1:T)}_{1:K+1} = \{\x^{(1:T)}_{1:K}, {\tilde{\y}}^{(1:T)}\}$.
\item In \textbf{phase~\rom{2}}, \generative and BERT-NER mutually improve each other iteratively in several loops.
Each loop first trains \generative with the observation $\x^{(1:T)}_{1:K+1}$ from the previous one.
Then, its predictions are adopted to fine-tune BERT-NER, whose output updates $\x^{(1:T)}_{K+1}$.
\end{itemize}
Figure~\ref{fig:3.cross.training} illustrates the alternate-training method.
In general, \generative gives high precision predictions, whereas BERT-NER trades recall with precision.
In other words, \generative can classify named entities with high accuracy but is slightly disadvantaged in discovering all entities.
BERT-NER increases the coverage with a certain loss of accuracy. 
Combined with the alternate-training approach, this complementarity between these models further increases the overall performance.

\begin{figure}[tbp]
    \centerline{\includegraphics[width = 0.48\textwidth]{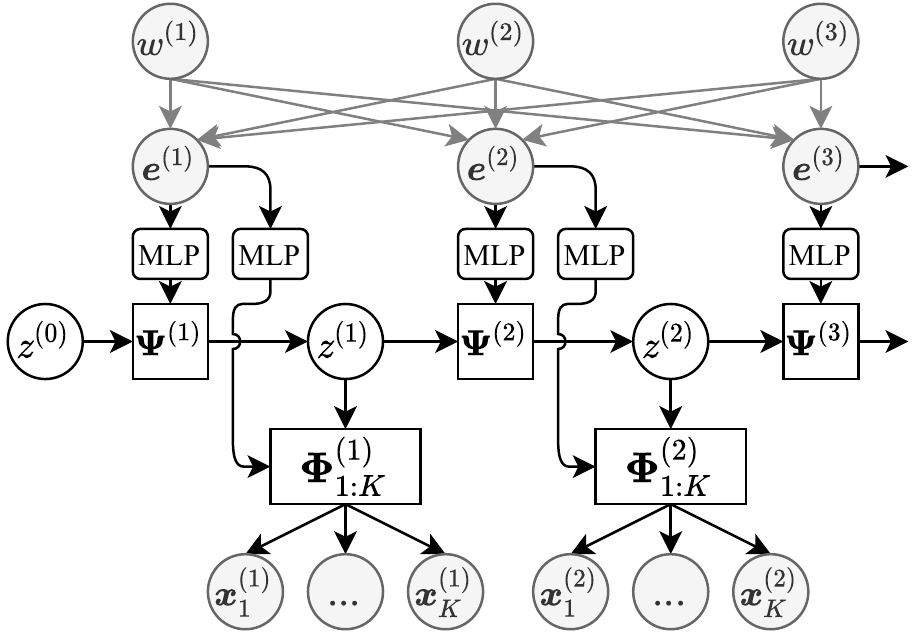}}
    \caption{
    An illustration of \generative's architecture.
    Shaded circles are observed elements; white circles are hidden elements; rectangles are matrices.
    Rounded rectangles are multi-layer perceptrons containing the trainable parameters.
    The arrows between $\w^{(t)}$ and $\e^{(t)}$ denote the context representation ability of BERT.
    MLP denotes the ``multi-layer perceptron''.
    }
    \label{fig:3-model-architecture}
\end{figure}

\subsection{Conditional Hidden Markov Model}
\label{subsec:method.nhmm}

The conditional hidden Markov model is an HMM variant for multi-source label denoising.
It models true entity labels as hidden variables and infers them from the observed noisy labels.
Traditionally, discrete HMM uses \textit{one} transition matrix to model the probability of hidden label transitioning and \textit{one} emission matrix to model the probability of the observations from the hidden labels.
These two matrices are constant, \ie, their values do not change over time steps.
\generative, on the contrary, conditions both its transition and emission matrices on the BERT embeddings $\e^{(1:T)}$ of the input tokens $\w^{(1:T)}$.
This design not only allows \generative to leverage the rich contextual representations of the BERT embeddings but relieves the constant matrices constraint as well.

In phase~\rom{1}, \generative takes $K$ sets of weak labels from the provided $K$ weak labeling sources.
In phase~\rom{2}, in addition to the existing sources, it takes another set of labels from the previously fine-tuned BERT-NER, making the total number of sources $K+1$.
For convenience, we use $K$ as the number of weak sources below.

\paragraph{Model Architecture}

Figure~\ref{fig:3-model-architecture} shows a sketch of \generative's architecture.\footnote{We relax plate notation here to present details.}
$\z^{(1:T)}$ denotes the discrete hidden states of \generative with $z^{(t)} \in \lbs$, representing the underlying true labels to be inferred from multiple weak annotations.
$\bPsi^{(t)} \in \R^{|\lbs| \times |\lbs|}$ is the transition matrix, whose element $\Psi^{(t)}_{i,j}=p(z^{(t)}=j|z^{(t-1)}=i, \e^{(t)}), i,j \in \{1,\dots,|\lbs|\}$ denotes the probability of moving from label $i$ to label $j$ at time step $t$.
$\bPhi^{(t)}_k \in \R^{|\lbs| \times |\lbs|}$ is the emission matrix of weak source $k$, each element in which $\Phi^{(t)}_{i,j,k}=p(x^{(t)}_{j,k}=1|z^{(t)}=i, \e^{(t)})$ represents the probability of source $k$ observing label $j$ when the hidden label is $i$ at time step $t$.

For each step, $\e^{(t)} \in \R^{d_{\rm emb}}$ is the output of a pre-trained BERT with $d_{\rm emb}$ being its embedding dimension.
$\bPsi^{(t)}$ and $\bPhi^{(t)}_{1:K}$ are calculated by applying a multi-layer perceptron (MLP) to $\e^{(t)}$:
\begin{gather}
    \label{eq:mlp-1}
    \bm{s}^{(t)} \in \R^{|\lbs|^2} = {\rm MLP}(\e^{(t)}), \\
    \label{eq:mlp-2}
    \bm{h}^{(t)} \in \R^{|\lbs|\cdot|\lbs|\cdot K} = {\rm MLP}(\e^{(t)}).
\end{gather}
Since the MLP outputs are vectors, we need to reshape them to matrices or tensors:
\begin{gather}
    \bm{S}^{(t)} \in \R^{|\lbs| \times |\lbs|} = {\rm reshape}(\bm{s}^{(t)}), \\
    \bm{H}^{(t)} \in \R^{|\lbs| \times |\lbs| \times K} = {\rm reshape}(\bm{h}^{(t)}).
\end{gather}
To achieve the proper probability distributions, we apply the Softmax function along the \textit{label} axis so that these values are positive and sum up to $1$:
\begin{equation*}
    \bPsi^{(t)}_{i,1:|\lbs|} = \sigma({\bm{S}}^{(t)}_{i,1:|\lbs|}), \ 
    \bPhi^{(t)}_{i,1:|\lbs|,k} = \sigma({\bm{H}}^{(t)}_{i,1:|\lbs|,k}),
\end{equation*}
where
\begin{equation}
    \sigma(\bm{a})_i = \frac{\exp{(a_i)}}{\sum_j \exp{(a_j)}}.
\end{equation}
$\bm{a}$ is an arbitrary vector.
The formulae in the following discussion always depend on $\e^{(1:T)}$, but we will omit the dependency term for simplicity.

\paragraph{Model Training}

According to the generative process of \generative, the joint distribution of the hidden states and the observed weak labels for one sequence $p(\z^{(0:T)}, \x^{(1:T)}|\btht)$ can be factorized as:
\begin{equation}
    \label{eq:hmm.joint.prob}
    \begin{aligned}
        & p(\z^{(0:T)}, \x^{(1:T)}|\btht) = p(z^{(0)})p(\x^{(1:T)}|\z^{(1:T)}) \\
        &\quad = p(z^{(0)})\prod_{t=1}^T p(z^{(t)}|z^{(t-1)}) \prod_{t=1}^T p(\x^{(t)}|z^{(t)}),
    \end{aligned}
\end{equation}
where $\btht$ represents all the trainable parameters.

HMM is generally trained with an expectation-maximization (EM, also known as Baum-Welch) algorithm.
In the expectation step (E-step), we compute the expected complete data log likelihood:
\begin{equation}
    \label{eq:em.q}
    Q(\btht, \btht^{\rm old}) \triangleq \expt_{\z} [\ell_c (\btht) | \btht^{\rm old}].
\end{equation}
$\btht^{\rm old}$ is the parameters from the previous training step, $\expt_{\z}[\cdot]$ is the expectation over variable $\z$, and
\begin{equation*}
    \ell_c(\btht) \triangleq \log p(\z^{(0:T)}, \x^{(1:T)} | \btht)
\end{equation*}
is the comptelete data log likelihood.
Let $\bphi^{(t)} \in \R^{|\lbs|}$ be the observation likelihood where
\begin{equation}
    \label{eq:obs.probs}
    \varphi^{(t)}_{i} \triangleq p(\x^{(t)}|z^{(t)}=i) = \prod_{k=1}^K\sum_{j=1}^{|\lbs|} \Phi^{(t)}_{i,j,k} x^{(t)}_{j,k}.
\end{equation}
Combining \eqref{eq:hmm.joint.prob}--\eqref{eq:obs.probs} together, we have
\begin{equation}
    \label{eq:q.calculate}
    \begin{aligned}
        & \ Q(\btht, \btht^{\rm old}) = \sum_{i=1}^{|\lbs|} \gamma^{(0)}_{i} \log \pi_i + \\
        & \sum_{t=1}^{T}\sum_{i=1}^{|\lbs|} \sum_{j=1}^{|\lbs|} \xi^{(t)}_{i,j} \log \Psi^{(t)}_{i,j}
        + \sum_{t=1}^{T}\sum_{i=1}^{|\lbs|} \gamma^{(t)}_{i} \log \varphi^{(t)}_{i},
    \end{aligned}
\end{equation}
where $\pi_1 = 1, \bm{\pi}_{2:|\lbs|} = \0$;\footnote{This assumes the initial hidden state is always \texttt{O}. In practice, we set $\pi_{\ell} = \epsilon, \forall \ell \in 2:|\lbs|$ and $\pi_{1}=1-(|\lbs|-1)\epsilon$, where $\epsilon$ is a small value, to avoid getting $-\infty$ from log.}
$\gamma^{(t)}_i \triangleq p(z^{(t)} = i| \x^{(1:T)})$ is the smoothed marginal; $\xi^{(t)}_{i,j} \triangleq p(z^{(t-1)}=i, z^{(t)}=j | \x^{(1:T)})$ is the expected number of transitions.
These parameters are computed using the \textit{forward-backward} algorithm.\footnote{Details are presented in \cref{appdseq:chmm.training}.}

In the maximization step (M-step), traditional HMM updates parameters ${\btht}_{\rm HMM} = \{ \bPsi, \bPhi, \bpi \}$ by optimizing \eqref{eq:em.q} with pseudo-statistics.\footnote{Details are presented in \cref{appdsubsec:hmm.mstep}.}
However, as the transitions and emissions in \generative are not standalone parameters, we cannot directly optimize \generative by this method.
Instead, we update the model parameters through gradient descent \wrt $\btht_{\rm \generative}$ using \eqref{eq:q.calculate} as the objective function:
\begin{equation}
    \label{eq:gradient}
    \nabla \btht_{\rm \generative} = \frac{\partial Q(\btht_{\rm \generative}, \btht_{\rm \generative}^{\rm old})}{\partial \btht_{\rm \generative}}.
\end{equation}

In practice, the calculation is conducted in the logarithm domain to avoid the loss of precision issue that occurs when the floating-point numbers become too small.

To solve the label sparsity issue, \ie, some entities are only observed by a minority of the weak sources, we modify the observations $\x^{(1:T)}$ before training.
If one source $k$ observes an entity at time step $t$: $x^{(t)}_{j\neq1, k}>0$, the observation of non-observing sources at $t$ will be modified to $\x^{(t)}_{1, \kappa}=\epsilon; \x^{(t)}_{j\neq1, \kappa}=(1-\epsilon)/|\lbs|, \forall \kappa \in \{1,\dots,K\}\backslash k$, where $\epsilon$ is an arbitrary small value.
Note that $x_{1, \kappa}^{(t)}$ corresponds to the observed label \texttt{O}.

\paragraph{CHMM Initialization}

Generally, HMM has its transition and emission probabilities initialized with the statistics $\bPsi^*$ and $\bPhi^*$ computed from the observation set.
But it is impossible to directly set $\bPsi^{(t)}$ and $\bPhi^{(t)}$ in \generative to these values, as these matrices are the output of the MLPs rather than standalone parameters.
To address this issue, we choose to pre-train the MLPs before starting \generative's training by minimizing the mean squared error (MSE) loss between their outputs and the target statistics:
\begin{equation*}
    \ell_{\rm MSE} = \frac{1}{T} \sum_t \| \bPsi^* - \bm{S}^{(t)} \|_F^2 + \| \bPhi^* - \bm{H}^{(t)} \|_F^2,
\end{equation*}
where $\|\cdot\|_F$ is the Frobenius norm.
Right after initialization, MLPs can only output similar probabilities for all time steps: $\bPsi^{(t)} \approx \bPsi^*,\ \bPhi^{(t)} \approx \bPhi^*,\ \forall t \in \{1,2,\dots,T\}$.
But their token-wise prediction divergence will emerge when \generative has been trained.
The initial hidden state $z^{(0)}$ is fixed to \texttt{O} as it has no corresponding token.

\paragraph{Inference}

Once trained, \generative can provide the most probable sequence of hidden labels $\hat{\z}^{(1:T)}$ along with the probabilities of all labels ${\y^*}^{(1:T)}$.
\begin{gather*}
    \hat{\z}^{(1:T)} = \argmax_{{\z}^{(1:T)}} p_{\hat{\btht}_{\rm \generative}}({\z}^{(1:T)} | \x^{(1:T)}_{1:K}, \e^{(1:T)}), \\
    {y^*}^{(t)}_i = p_{\hat{\btht}_{\rm \generative}}(z^{(t)}=i | \x^{(1:T)}_{1:K}, \e^{(1:T)}),
\end{gather*}
where $\hat{\btht}_{\rm \generative}$ represents the trained parameters.
These results can be calculated by either the Viterbi decoding algorithm \citep{viterbi-1967-error} or directly maximizing the smoothed marginal $\bm{\gamma}^{(1:T)}$.


\subsection{Improving Denoised Labels with BERT}
\label{subsec:bert.ner}

The pre-trained BERT model encodes semantic and structural knowledge, which can be distilled to further refine the denoised labels from \generative.
Specifically, we construct the BERT-NER model by stacking a feed-forward layer and a Softmax layer on top of the original BERT to predict the probabilities of the classes that each token belongs to \citep{sun-2019-fine-tune}.
The probability predictions of \generative, ${\y^*}^{(1:T)}$, often referred to as \textit{soft labels}, are chosen to supervise the fine-tuning procedure.
Compared with the hard labels $\hat{\z}^{(1:T)}$, soft labels lead to a more stable training process and higher model robustness \citep{Thiel-2008-classification, Liang-etal-2020-bond}.

We train BERT-NER by minimizing the Kullback-Leibler divergence (KL divergence) between the soft labels $\y^*$ and the model output $\y$:
\begin{equation}
    \begin{aligned}
        \hat{\btht}_{\rm BERT} &= \argmin_{{\btht}_{\rm BERT}} \kld[{\y^*}^{(1:T)} \| \y^{(1:T)} ] \\
        &= \argmin_{{\btht}_{\rm BERT}} \sum_{t=1}^T\sum_{i=1}^{|\lbs|} {y^*}^{(t)}_i \log \frac{{y^*}^{(t)}_i}{{y}_i^{(t)}},
    \end{aligned}
\end{equation}
where ${\btht}_{\rm BERT}$ denotes all the trainable parameters in the BERT model.
BERT-NER does not update the embeddings $\e^{(1:T)}$ that \generative depends on.

We obtain the refined labels ${\tilde{\y}}^{(1:T)} \in \R^{T \times |\lbs|}$ from the fine-tuned BERT-NER directly through a forward pass.
Different from \generative, we continue BERT-NER's training with parameter weights from the last loop's checkpoint so that the model is initialized closer to the optimum.
Correspondingly, phase~\rom{2} trains BERT-NER with a smaller learning rate, fewer epoch iterations, and batch gradient descent instead of the mini-batch version.\footnote{Hyper-parameter values are listed in \cref{appdsec:hyper.parameters}.}
This strategy speeds up phase~\rom{2} training without sacrificing the model performance as ${\y^*}^{(1:T)}$ does not change significantly from loop to loop.

\section{Experiments}
\label{sec:experiments}

We benchmark \ours on four datasets against state-of-the-art weakly supervised NER baselines, including both distant learning models and multi-source label aggregation models.
We also conduct a series of ablation studies to evaluate the different components in \ours's design.

\begin{table}[htbp]\small
    \centering
    \begin{tabular}{c|c|c|c|c}
    \toprule
     & Co03 & NCBI & CDR & LR \\
    \midrule
    \# Instance & \num{22137} & \num{793} & \num{1500} & \num{3845} \\
    \# Training & \num{14041} & \num{593} & \num{500} & \num{2436} \\
    \# Development & \num{3250} & \num{100} & \num{500} & \num{609} \\
    \# Test & \num{3453} & \num{100} & \num{500} & \num{800} \\
    \midrule
    Ave\# Tokens & \num{14.5} & \num{219.8} & \num{217.7} & \num{16.4} \\
    \midrule
    \# Entities & \num{4} & \num{1} & \num{2} & \num{1} \\
    \# Sources & \num{13} & \num{5} & \num{8} & \num{4} \\
    \bottomrule
    \end{tabular}
    \caption{
    Dataset statistics.
    Co03 is CoNLL 2003; LR is LaptopReview; CDR is BC5CDR.
    ``\# Sources'' indicates the number of labeling sources for each dataset.
    }
    \label{tb:dataset.statistics}
\end{table}

\subsection{Setup}
\label{subsec:exp.setup}

\paragraph{Datasets}
We consider four NER datasets covering the general, technological and biomedical domains:
1) \textbf{CoNLL 2003} (English subset) \citep{tjong-kim-sang-de-meulder-2003-introduction} is a general domain dataset containing \num{22137} sentences manually labelled with \num{4} entity types.
2) \textbf{LaptopReview} dataset \citep{pontiki-etal-2014-semeval} consists of \num{3845} sentences with laptop-related entity mentions.
3) \textbf{NCBI-Disease} dataset \citep{Dogan-2014-NCBI} contains \num{793} PubMed abstracts annotated with disease mentions.
4) \textbf{BC5CDR} \citep{Li-2016-BC5CDR}, the dataset accompanies the BioCreative V CDR challenge, consists of \num{1500} PubMed articles, annotated with chemical disease mentions.

Table~\ref{tb:dataset.statistics} shows dataset statistics, including the average number of tokens, entities and weak labeling sources.
We use the original word tokens in the dataset if provided and use NLTK \citep{bird-loper-2004-nltk} otherwise for sentence tokenization.

For weak labeling sources, we use the ones from \citet{lison-etal-2020-named} for CoNLL 2003, and the ones from \citet{Safranchik-etal-2020-weakly} for LaptopReview, NCBI-Disease and BC5CDR.\footnote{Details are presented in \cref{appdsec:labeling.source.performance}.}

\begin{table*}[tbp]\small
    \centering
    \begin{tabular}{c|c|c|c|c}
    \toprule
    Models & CoNLL 2003 & NCBI-Disease & BC5CDR & LaptopReview \\
    \midrule
    Supervised BERT-NER $\ddagger$ $\natural$ & 90.74 (90.37/91.10) & 88.89 (87.05/90.82) & 88.81 (87.12/90.57) & 81.34 (82.02/80.67)  \\
    best consensus $\natural$ & 89.18 (100.0/80.47) & 81.60 (100.0/68.91) & 87.58 (100.0/77.89) & 77.72 (100.0/63.55) \\
    \midrule
    SwellShark (noun-phrase) $\dagger \ddagger$ & - & 67.10 (64.70/69.70) & 84.23 (84.98/83.49) & -  \\
    SwellShark (hand-tuned) $\dagger \ddagger$ & - & 80.80 (81.60/80.10) & 84.21 (86.11/82.39) & - \\
    AutoNER $\dagger \ddagger$ & 67.00 (75.21/60.40) & 75.52 (79.42/71.98) & 82.13 (83.23/81.06) & 65.44 (72.27/59.79) \\
    Snorkel $\dagger \ddagger$ & 66.40 (71.40/62.10) & 73.41 (71.10/76.00) & 82.24 (80.23/84.35) & 63.54 (64.09/63.09) \\
    Linked HMM $\dagger \ddagger$ & - & 79.03 (83.46/75.05) & 82.96 (82.65/83.28) & 69.04 (77.74/62.11) \\
    BOND-MV $\dagger \ddagger$ $\natural$ &65.96 (64.22/67.82) & 80.33 (84.77/76.34) & 83.18 (82.90/83.49) & 67.19 (68.90/65.75) \\
    \midrule
    Majority Voting $\dagger$ $\natural$  & 58.40 (49.01/72.24) & 73.94 (79.76/68.91) & 80.73 (83.79/77.88) & 67.92 (72.93/63.55)  \\
    HMM $\dagger$ $\natural$  & 68.84 (70.80/66.98) & 73.06 (83.88/64.70)  & 80.57 (88.75/73.76) & 66.96 (77.46/58.96) \\
    \generative-i.i.d. $\dagger$ $\natural$ & 68.57 (69.67/67.50) & 71.69 (83.49/62.87) & 79.37 (85.68/73.92) & 65.89 (75.70/58.34)  \\
    \midrule
    \generative $\dagger$ $\natural$ & 70.11 (72.98/67.47) & 78.88 (\textbf{93.37}/68.28) & 82.39 (\textbf{89.93}/76.02) &  73.02 (\textbf{87.23}/62.79) \\
    \generative + BERT-NER $\dagger \ddagger$ $\natural$ & 74.30 (75.02/73.58) & 82.87 (89.42/77.22) & 84.33 (85.58/83.12) & 69.67 (75.48/64.70) \\
    \ours $\dagger \ddagger$ $\natural$ & \textbf{75.54} (\textbf{76.22}/\textbf{74.86}) & \textbf{85.02} (87.92/\textbf{82.47}) & \textbf{85.12} (84.97/\textbf{85.28}) & \textbf{76.55} (81.39/\textbf{72.32}) \\
    \bottomrule
    \end{tabular}
    \caption{
        Evaluation results on four datasets.
        The results are presented in the ``F1 (Precision/Recall)'' format.
        ``\generative+ BERT-NER'' is essentially \ours's phase~\rom{1} output.
        ``BOND-MV'' is the BOND model trained with majority voted labels.
        $\dagger$ indicates unsupervised label denoiser; $\ddagger$ represents fully supervised models.
        A model with $\dagger \ddagger$ is either distantly supervised or trains a supervised by labels from the denoiser.
        $\natural$ signifies the results from our experiments.
        In addition to models with $\natural$, Snorkel and Linked HMM also share our labeling sources.
    }
    \label{tb:results.domains}
\end{table*}

\paragraph{Baselines}
We compare our model to the following state-of-the-art baselines:
1)~\textbf{Majority Voting} returns the label for a token that has been observed by most of the sources and randomly chooses one if it's a tie;
2)~\textbf{Snorkel} \citep{Ratner-2017-Snorkel} treats each token in a sequence as i.i.d. and conducts the label classification without considering its context;
3)~\textbf{SwellShark} \citep{Fries-2017-SwellShark} improves Snorkel by predicting all the target entity spans before classifying them using na\"ive Bayes;
4)~\textbf{AutoNER} \citep{shang-etal-2018-learning} augments distant supervision by predicting whether two consecutive tokens should be in the same entity span;
5)~\textbf{BOND} \citep{Liang-etal-2020-bond} adopts self-training and high-confidence selection to further boost the distant supervision performance.
6)~\textbf{HMM} is the multi-observation generative model used in \citet{lison-etal-2020-named} that does not have the integrated neural network;
7)~\textbf{Linked HMM} \citep{Safranchik-etal-2020-weakly} uses linking rules to provide additional inter-token structural information to the HMM model.

For the ablation study, we modify \generative to another type of i.i.d. model by taking away its transition matrices.
This model, named \textbf{\generative-i.i.d.}, directly predicts the hidden steps from the BERT embeddings, while otherwise identical to \generative.
We also investigate how \ours performs with other aggregators other than \generative.

We also introduce two upper bounds from different aspects:
1) a \textbf{fully supervised BERT-NER} model trained with manually labeled data is regarded as a supervised reference;
2) the \textbf{best possible consensus} of the weak sources.
The latter assumes an oracle that always selects the correct annotations from these weak supervision sources.
According to the definition, its precision is always $100\%$ and its recall is non-decreasing with the increase of the number of weak sources.

\paragraph{Evaluation Metrics}

We evaluate the performance of NER models using entity-level precision, recall, and F1 scores.
All scores are presented as percentages.
The results come from the average of \num{5} trials with different random seeds.

\paragraph{Implementation Details}

We use BERT pre-trained on different domains for different datasets, both for embedding construction and as the component of the supervised BERT-NER model.
The original BERT \citep{devlin-etal-2019-bert} is used for CoNLL 2003 and LaptopReview datasets, bioBERT \citep{Lee-2019-biobert} for NCBI-Disease and SciBERT \citep{beltagy-etal-2019-scibert} for BC5CDR.
Instances with lengths exceeding BERT's maximum length limitation (\num{512}) are broken into several shorter segments.

The only tunable hyper-parameter in \generative is the learning rate.
But its influence is negligible---benefitted from the stability of the generalized EM, the model is guaranteed to converge to a local optimum if the learning rate is small enough.
For all the BERT-NER models used in our experiments, the hyper-parameters except the batch size are fixed to the default values (\cref{appdsec:hyper.parameters}).

To prevent overfitting, we use a two-scale early stopping strategy for model choosing at two scales based on the development set.
The micro-scale early stopping chooses the best model parameters for each individual training process of both \generative and BERT-NER;
the macro-scale early stopping selects the best-performing model in phase~\rom{2} iterations, which reports the test results.
In our experiments, phase~\rom{2} exits if the macro-scale development score has not increased in \num{5} loops or the maximum number of loops (\num{10}) is reached.

\begin{figure*}[tbp]
    \centering {
        \subfloat[] {
            \label{subfig:n-ted}
            \includegraphics[width=0.235\textwidth]{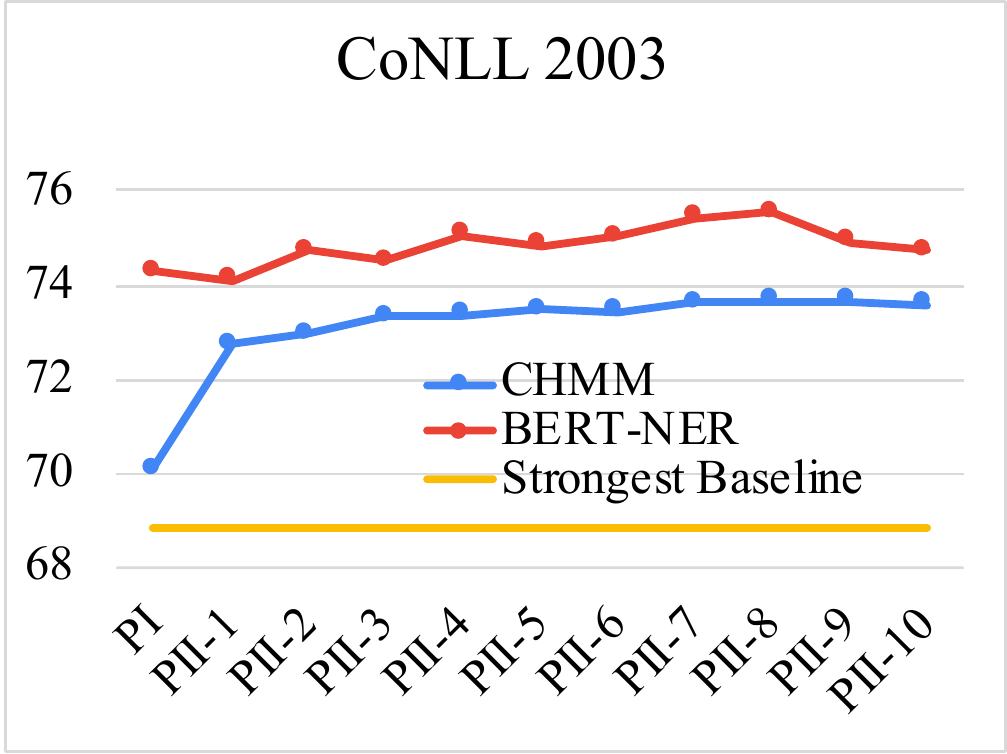}
        }
        \subfloat[] {
            \label{subfig:ted}
            \includegraphics[width=0.235\textwidth]{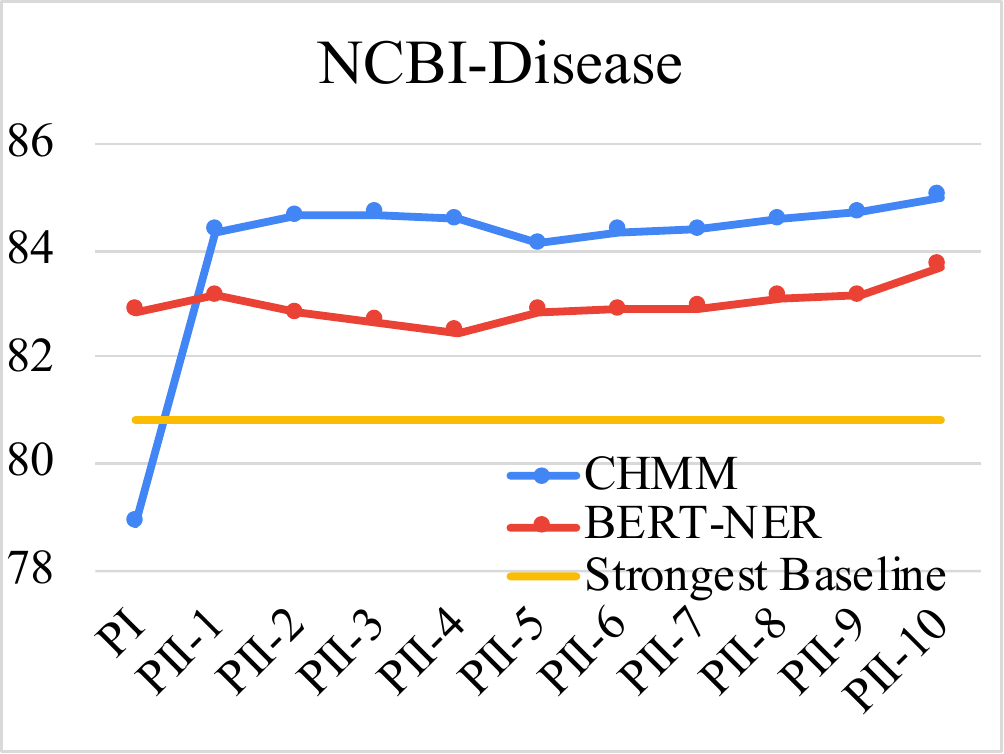}
        }
        \subfloat[] {
            \label{subfig:n-ted}
            \includegraphics[width=0.235\textwidth]{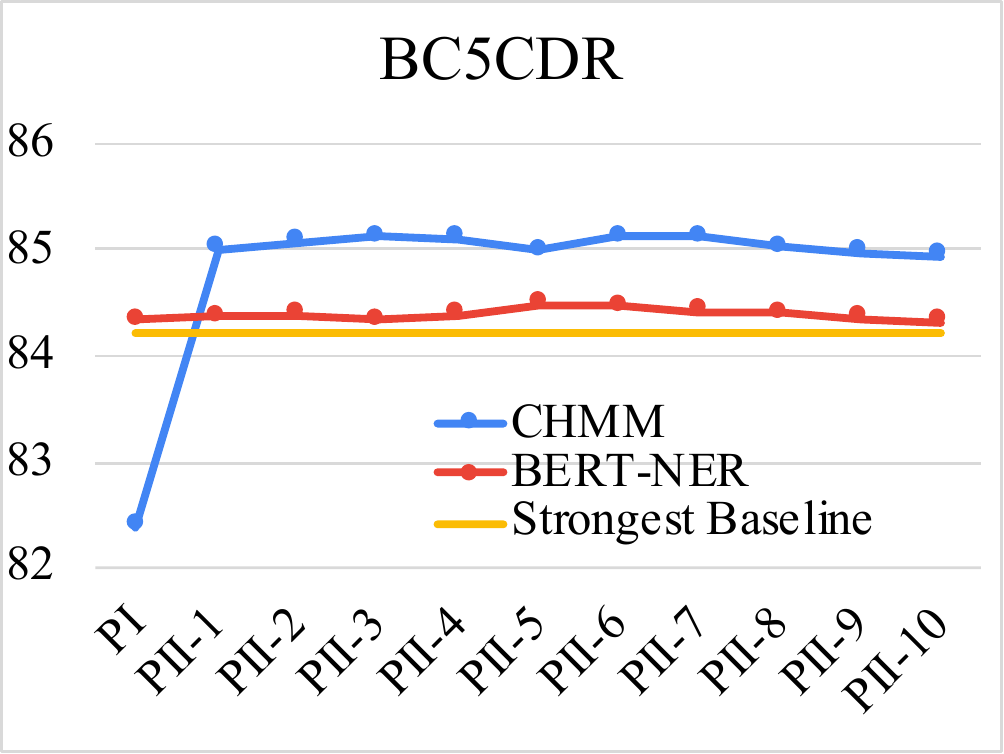}
        }
        \subfloat[] {
            \label{subfig:n-ted}
            \includegraphics[width=0.235\textwidth]{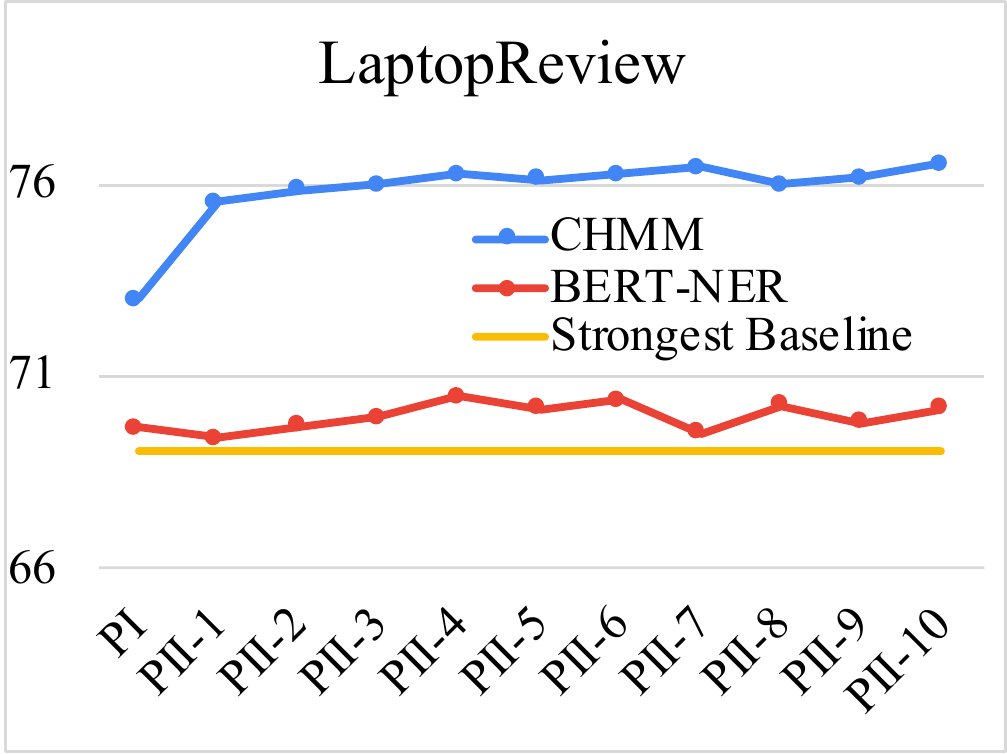}
        }
    }
    \caption{
        F1 score evolution across the alternate-training phases.
        ``PI'' is phase~\rom{1}; ``PII-$i$'' is the $i$th loop of phase~\rom{2}.
        The ``strongest baseline'' reports the result from the best-performed baseline in Table~\ref{tb:results.domains} for each dataset.
    }
    \label{fig:cross.training.performance}
\end{figure*}

\subsection{Main Results}

Table~\ref{tb:results.domains} presents the model performance from different domains.
We find that our alternate-training framework outperforms all weakly supervised baseline models.
In addition, \ours approaches or even exceeds the best source consensus, which sufficiently proves the effectiveness of the design.
For general HMM-based label aggregators such as \generative, it is impossible to exceed the best consensus since they can only predict an entity observed by at least one source.
Based on this fact, \generative is designed to select the most accurate observations from the weak sources without shrinking their coverage.
In comparison, BERT's language representation ability enables it to generalize the entity patterns and successfully discovers those entities annotated by none of the sources.
Comparing \generative + BERT to \generative, we can conclude that BERT basically exchanges recall with precision, and its high-recall predictions can improve the result of \generative in return.
The complementary nature of these two models is why \ours improves the overall performance of weakly supervised NER.


\subsection{Analysis of CHMM}

Looking at Table~\ref{tb:results.domains}, we notice that \generative performs the best amongst all generative models including majority voting, HMM and \generative-i.i.d.
The performance of conventional HMM is largely limited by the Markov assumption with the unchanging transition and emission probabilities.
The results in the table validate that conditioning the model on BERT embedding alleviates this limitation.
However, the transition matrices in HMM are indispensable, implied by \generative-i.i.d.'s results, as they provide supplemental information about how the underlying true labels should evolve.

\begin{table}[tbp]\small
    \centering
    \begin{tabular}{c|cccc}
    \toprule
    Models        & Co03 & NCBI  & CDR & Laptop  \\
    \midrule
    MV $\dagger$ $\natural$ &  58.40 & 73.94  & 80.73 & 67.92 \\
    MV-ALT $\dagger \ddagger$ $\natural$ & 66.64 & 80.83 & 82.78  & 70.45  \\
    \midrule
    HMM $\dagger$ $\natural$ & 68.84  & 73.06 & 80.57 & 66.96 \\
    HMM-ALT $\dagger \ddagger$ $\natural$          & 74.04  & 82.99 & 83.34  & 72.90      \\
    \midrule
    i.i.d. $\dagger$ $\natural$ & 68.57 & 71.69 & 79.37 & 65.89 \\
    i.i.d.-ALT $\dagger \ddagger$ $\natural$      & 73.84 & 83.15 & 83.17  & 72.61       \\
    \midrule
    \generative $\dagger$ $\natural$ & 70.11 & 78.88 & 82.39 &  73.02 \\
    \ours $\dagger \ddagger$ $\natural$        & 75.54   & 85.02 & 85.12  & 76.55\\
    \bottomrule
    \end{tabular}
    \caption{
        Alternate-training F1 scores with different label aggregators.
        MV denotes Majority voting; i.i.d. represents \generative-i.i.d.
        The model names without the ``ALT'' suffix are the multi-source label aggregators whereas the suffix indicates that the result comes from the alternate-training framework with the corresponding model as the label aggregator.
    }
    \label{tb:results.cross.training}
\end{table}

\subsection{Analysis of Alternate-Training}

\paragraph{Performance Evolution}

Figure~\ref{fig:cross.training.performance} reveals the details of the alternate-training process.
For less ambiguous tasks including NCBI-Disease, BC5CDR and LaptopReview with fewer entity types, BERT generally has better performance in phase~\rom{1} but gets surpassed in phase~\rom{2}.
Interestingly, BERT's performance never exceeds that of \generative on the LaptopReview dataset.
This may be because BERT fails to construct sufficiently representative patterns from the denoised labels for this dataset.
For CoNLL 2003, where it is harder for the labeling sources to model the language structures, the strength of a pre-trained language model in pattern recognition becomes more prominent.
From the results it seems that the performance increment of the denoised labels ${\y^*}^{(1:T)}$ provides marginally extra information to BERT after phase~\rom{2}, as most of the increment comes from the information provided by BERT itself.
Even so, keeping phase~\rom{2} is reasonable when we want to get the best out of the weak labeling sources and the pre-trained BERT.

\paragraph{BERT-NER Initialization}

\ours initializes BERT-NER's parameters from its previous checkpoint at the beginning of each loop in phase~\rom{2} to reduce training time (\cref{subsec:bert.ner}).
If we instead fine-tune BERT-NER from the initial parameters of the pre-trained BERT model for each loop, \ours gets \num{84.30}, \num{84.71}, and \num{76.68} F1 scores on NCBI-Disease, BC5CDR, and LaptopReview datasets.
These scores are close to the results in Table~\ref{tb:results.domains}, but the training takes much longer.
Consequently, our BERT-NER initialization strategy is a more practical choice overall.

\paragraph{Applying Alternate-Training to Other Methods}

Table~\ref{tb:results.cross.training} shows the alternate-training performance acquired with different label aggregators.
The accompanying BERT-NER models are identical to those described in \cref{subsec:exp.setup}.
The results in the table suggest that the performance improvement obtained by using alternate-training on the label aggregators is stable and generalizable to any other models yet to be proposed.

\section{Conclusion}
\label{sec:conclusion}

In this work, we present \ours, a multi-source weakly supervised approach that does not depend on manually labeled data to learn an accurate NER tagger.
It integrates a label aggregator---\generative and a supervised model---BERT-NER together into an alternate-training procedure.
\generative conditions HMM on BERT embeddings to achieve greater flexibility and stronger context-awareness.
Fine-tuned with \generative's prediction, BERT-NER discovers patterns unobserved by the weak sources and complements \generative.
Training these models in turn, \ours uses the knowledge encoded in both the weak sources and the pre-trained BERT model to improve the final NER performance.
In the future, we will consider imposing more constraints on the transition and emission probabilities, or manipulating them according to sophisticated domain knowledge.
This technique could be also extended to other sequence labeling tasks such as semantic role labeling or event extraction.

\section*{Acknowledgments}

This work was supported by ONR MURI N00014-17-1-2656, NSF III-2008334, Kolon Industries, and research gifts from Google and Amazon.
In addition, we would like to thank Yue Yu for his insightful suggestions for this work.

\bibliographystyle{acl_natbib}
\bibliography{references}

\clearpage

\appendix

\section{Technical Details}
\label{appdseq:technical.details}

\subsection{\generative Training}
\label{appdseq:chmm.training}

Following the discussion in \cref{subsec:method.nhmm}, we use the \textit{forward-backward} algorithm to calculate the smoothed marginal $\gamma^{(t)}_i \triangleq p(z^{(t)} = i| \x^{(1:T)}), i \in \{1, 2, \dots, |\lbs| \}, t \in \{1, 2, \dots, T \}$ and the expected number of transitions $\xi^{(t)}_{i,j} \triangleq p(z^{(t-1)}=i, z^{(t)}=j | \x^{(1:T)}), i,j \in \{1, 2, \dots, |\lbs| \}$.\footnote{Same as \cref{subsec:method.nhmm}, we omit the dependency term $\e^{(1:T)}$.}
$|\lbs|$ is the number of BIO formatted entity labels, which are regarded as hidden states;
$T$ is the total number of hidden steps in a sequence, which equals the number of tokens.

Defining $\alpha^{(t)}_i \triangleq p(z^{(t)}=i|\x^{(1:t)})$ and $\beta^{(t)}_i \triangleq p(\x^{(t+1:T)}|z^{(t)} = i)$, $\gamma^{(t)}_i$ and $\xi^{(t)}_{i,j}$ can be represented by $\alpha$ and $\beta$ using the Bayes' rule and Markov assumption:
\begin{gather}
    \label{appdeq:gamma.cal.scalar}
    \begin{aligned}
        \gamma^{(t)}_i &\triangleq p(z^{(t)} = i| \x^{(1:T)}) \\
        &= \frac{p(\x^{(t+1:T)}, z^{(t)} = i | \x^{(1:t)})}{p(\x^{(t+1:T)}|\x^{(1:T)})} \\
        &\propto p(z^{(t)}=i|\x^{(1:t)}) p(\x^{(t+1:T)}|z^{(t)} = i) \\
        &= \alpha^{(t)}_i \beta^{(t)}_i,
    \end{aligned} \\
    \label{appdeq:xi.cal.scalar}
    \begin{aligned}
        \xi^{(t)}_{i,j} &\triangleq p(z^{(t-1)}=i, z^{(t)}=j | \x^{(1:T)}) \\
        & \propto p(z^{(t-1)}=i|\x^{(1:t-1)}) \\
        & \quad \ p(z^{(t)}=j|z^{(t-1)}=i, \x^{(t:T)}) \\
        & \propto p(z^{(t-1)}=i|\x^{(1:t-1)}) p(\x^{(t)}|z^{(t)}=j) \\
        & \quad \ p(\x^{(t+1:T)}|z^{(t)}=j)p(z^{(t)}=j|z^{(t-1)}=i)\\
        &= \alpha^{(t-1)}_i\varphi^{(t)}_j\beta^{(t)}_j\Psi^{(t)}_{i,j}.
      \end{aligned}
\end{gather}
$\varphi^{(t)}_i \in \R^{|\lbs|} \triangleq p(\x^{(t)}|z^{(t)}=i)$ is the likelihood of the observation when the hidden state is $i$ (\cref{subsec:method.nhmm}).

Written in the matrix form, \eqref{appdeq:gamma.cal.scalar} and \eqref{appdeq:xi.cal.scalar} become:
\begin{gather}
    \bm{\gamma}^{(t)} \propto \ba^{(t)} \odot \bm{\beta}^{(t)}, \\
    \bm{\xi}^{(t)} \propto \bPsi^{(t)} \odot (\ba^{(t-1)}(\bphi^{(t)} \odot \bm{\beta}^{(t)})^{\sf T}),
\end{gather}
where $\odot$ is the element-wise product.
Note that the elements in both $\bm{\gamma}^{(t)}$ and $\bm{\xi}^{(t)}$ should sum up to $1$.

\paragraph{The Forward Pass}

The filtered marginal $\alpha^{(t)}_i$ can be computed iteratively:
\begin{equation}
  \label{appdeq:forward}
  \begin{aligned}
    \alpha^{(t)}_i &\triangleq p(z^{(t)}=i|\x^{(1:t)}) \\
    &= p(z^{(t)}=i|\x^{(t)}, \x^{(1:t-1)}) \\
    &\propto p(\x^{(t)}|z^{(t)}=i) p(z^{(t)}=i|\x^{(1:t-1)}) \\
    &= \sum_j \varphi^{(t)}_i \Psi^{(t)}_{j,i} \alpha^{(t-1)}_j.
  \end{aligned}
\end{equation}
Written in the matrix form, \eqref{appdeq:forward} becomes
\begin{equation}
    \ba^{(t)} \propto \bphi^{(t)} \odot ({\bPsi^{(t)}}^{\sf T}\ba^{(t-1)}).
\end{equation}
We initialize $\ba$ with $\ba^{(0)} = \bpi$ (\cref{subsec:method.nhmm}) since we have no observation at time step $0$.
As $\ba^{(t)}$ is a probability distribution, the elements in it sum up to $1$.
The calculation of $\ba$ is the \textit{forward pass}.

\paragraph{The Backward Pass}

In the same way, we do the \textit{backward pass} to compute the conditional future evidence $\beta^{(t)}_i \triangleq p(\x^{(t+1:T)}|z^{(t)} = i)$:
\begin{equation}
    \label{appdeq:backward}
    \begin{aligned}
      \beta^{(t-1)}_i &\triangleq p(\x^{(t+1:T)}|z^{(t)} = j) \\
      &= \sum_j p(z^{(t)}=j, \x^{(t)}, \x^{(t+1:T)}|z^{(t-1)}=i) \\
      &= \sum_j [p(\x^{(t+1:T)}|z^{(t)}=j) \\ &\qquad \quad p(\x^{(t)}, z^{(t)} = j|z^{(t-1)}=i)] \\
      &= \sum_j \beta^{(t)}_j \varphi^{(t)}_j \Psi^{(t)}_{i,j}.
    \end{aligned}
\end{equation}
In the matrix form, \eqref{appdeq:backward} becomes:
\begin{equation}
    \bm{\beta}^{(t-1)} = \bPsi^{(t)}(\bphi^{(t)} \odot \bm{\beta}^{(t)}),
\end{equation}
whose base case is
\begin{equation*}
    \begin{aligned}
        \beta^{(T)}_i = p(\x^{(T+1:T)}|z^{(T)}=i) &= 1, \\
        \forall i & \in \{1, \dots, |\lbs|\}.
    \end{aligned}
\end{equation*}

\subsection{The Maximization step for Unsupervised HMM}
\label{appdsubsec:hmm.mstep}

For traditional unsupervised HMM, the expected complete data log likelihood is maximized by updating the matrices with the approximated pseudo-statistics.
different from \generative, HMM has constant transition and emission for all time steps, \ie:
\begin{equation*}
    \begin{aligned}
        \bPsi^{(1)} = \bPsi^{(t)};
        \bPhi^{(1)} = \bPhi^{(t)};
        \quad \forall t \in \{2, \dots, T \}.
    \end{aligned}
\end{equation*}
For simplicity, we remove the term $t$ for the transition and emission matrices.
Suppose we are updating HMM based on one instance with $t$ starting from $1$:
\begin{gather}
  \pi_i = \gamma^{(1)}_i; \\
  \Psi_{i, j} = \frac{\sum_{t=2}^{T}\xi^{(t)}_{i,j}}{\sum_{t=2}^{T}\sum_{\ell=1}^{|\lbs|}\xi^{(t)}_{i,\ell}}; \\
  \Phi_{i, j, k} = \frac{\sum_{t=1}^{T}\gamma^{(t)}_i x^{(t)}_{j, k}} {\sum_{t=1}^{T}\gamma^{{t}}_i}.
\end{gather}
Note that the observation has property $0 \leq x^{(t)}_{j, k} \leq 1$ and $\sum_{j=1}^{|\lbs|} x^{(t)}_{j, k}=1$, where $k\in\{1, \dots K\}$ is the index of the weak labeling source.

\section{Labeling Source Performance}
\label{appdsec:labeling.source.performance}

The weak labeling sources of the CoNLL 2003 dataset come from \citet{lison-etal-2020-named}, whereas \citet{Safranchik-etal-2020-weakly} provide the sources for the LaptopReview, NCBI-Disease and BC5CDR dataset.
For \citet{Safranchik-etal-2020-weakly}'s labeling sources, we apply a majority voting using their tagging results to the spans detected by their \textit{linking rules} to convert the linking results to token annotations.
In consideration of the training time and resource consumption, we only adopt a subset of the labeling sources provided by the authors.
The performance of the labeling sources is presented in the tables below.

\begin{table}[htbp]\small
    \centering
    \begin{tabular}{c|ccc}
    \toprule
    source name                  & precision & recall & f1     \\
    \midrule
    CoreDictionaryUncased & 81.03    & 41.41 & 5.48  \\
    CoreDictionaryExact   & 80.69    & 17.18 & 28.32 \\
    CancerLike            & 34.88    & 1.58 & 3.02 \\
    BodyTerms             & 68.52    & 3.90  & 7.38 \\
    ExtractedPhrase       & 97.12    & 32.03 & 48.18 \\
    \bottomrule
    \end{tabular}
    \caption{The performance of the labeling sources used in the NCBI-Disease dataset.}
\end{table}

\begin{table}[htbp]\small
    \centering
    \begin{tabular}{c|ccc}
    \toprule
    source name                  & precision & recall & f1     \\
    \midrule
    DictCore-Chemical       & 91.81    & 29.55 & 44.7  \\
    DictCore-Chemical-Exact & 85.88    & 3.16 & 6.1  \\
    DictCore-Disease        & 81.57    & 26.32 & 39.8  \\
    DictCore-Disease-Exact  & 81.4     & 1.09 & 2.16 \\
    Organic Chemical        & 92.67    & 30.07 & 45.4  \\
    Disease or Syndrome     & 77.36    & 11.67 & 20.28 \\
    PostHyphen              & 84.47    & 08.07 & 14.74 \\
    ExtractedPhrase         & 86.8     & 17.96 & 29.76 \\
    \bottomrule
    \end{tabular}
    \caption{The performance of the labeling sources used in the BC5CDR dataset.}
\end{table}

\begin{table}[htbp]\small
    \centering
    \begin{tabular}{c|ccc}
    \toprule
    source name                  & precision & recall & f1     \\
    \midrule
    CoreDictionary      & 72.63    & 51.61 & 60.34 \\
    iStuff              & 26.67    & 0.61 & 1.2  \\
    ExtractedPhrase     & 97.45    & 29.25 & 45.0  \\
    ConsecutiveCapitals & 35.29    & 0.92 & 1.8  \\
    \bottomrule
    \end{tabular}
    \caption{The performance of the labeling sources used in the LaptopReview dataset.}
\end{table}

\begin{table}[htbp]\small
    \centering
    \begin{tabular}{c|ccc}
    \toprule
    source name                  & precision & recall & f1     \\
    \midrule
    BTC+c                  & 61.56 & 46.35 & 52.88 \\
    SEC+c                  & 39.54 & 24.59 & 30.32 \\
    core\_web\_md+c        & 69.53 & 60.04 & 64.44 \\
    crunchbase\_cased      & 38.26 & 5.59 & 9.76 \\
    crunchbase\_uncased    & 37.88 & 6.2  & 10.66 \\
    doc\_majority\_cased   & 65.81 & 40.21 & 49.92 \\
    doc\_majority\_uncased & 61.69 & 40.17 & 48.66 \\
    full\_name\_detector   & 87.79 & 11.33 & 20.06 \\
    geo\_cased             & 68.16 & 15.35 & 25.06 \\
    geo\_uncased           & 65.1  & 18.89 & 29.28 \\
    misc\_detector         & 85.14 & 21.51 & 34.34 \\
    wiki\_cased            & 75.27 & 32.65 & 45.54 \\
    wiki\_uncased          & 72.26 & 35.61 & 47.7  \\
    \bottomrule
    \end{tabular}
    \caption{The performance of the labeling sources used in the CoNLL 2003 dataset.}
\end{table}

Please refer to \citet{lison-etal-2020-named} for the information about the construction of the labeling sources on the CoNLL 2003 dataset;
please refer to \citet{Safranchik-etal-2020-weakly} for the labeling sources on other three datasets.

\section{Hyper-Parameters}
\label{appdsec:hyper.parameters}

The experiments are conducted on one GeForce RTX 2080 Ti GPU.
For NCBI-Disease, BC5CDR and LaptopReview datasets, \generative is pre-trained for \num{5} epochs and trained for \num{20} epochs.
The learning rates for these three datasets are $5\times 10^{-4}$, $10^{-3}$ and $10^{-4}$, respectively, and the batch sizes are \num{64}, \num{64} and \num{128}.
In phase~\rom{1}, BERT-NER is trained with the default learning rate ($5\times 10^{-5}$) for $100$ epochs.
The batch sizes are $8$, $8$, and $48$, respectively.
Note that for LaptopReview, the maximum length limitation of BERT-NER is set to \num{128} whereas the limitation is \num{512} for the other two datasets.
In phase~\rom{2}, we use half the learning rate with $20$ epochs for each loop.

For CoNLL 2003, \generative has the same number of training epochs as for other datasets.
The batch size is $32$, and the learning rate is $10^{-5}$.
BERT-NER has a maximum sequence length of $256$.
It is trained for $15$ epochs in phase~\rom{1} and $5$ epochs in phase~\rom{2}.
Other hyper-parameters are identical to other BERT-NER models'.

\end{document}